# The Road to Quantum Artificial Intelligence

**Kyriakos N. Sgarbas**

Wire Communications Lab., Dept. of Electrical and Computer Engineering,
University of Patras, GR-26500, Patras, Greece
E-mail: sgarbas@upatras.gr

## Abstract

This paper overviews the basic principles and recent advances in the emerging field of Quantum Computation (QC), highlighting its potential application to Artificial Intelligence (AI). The paper provides a very brief introduction to basic QC issues like quantum registers, quantum gates and quantum algorithms and then it presents references, ideas and research guidelines on how QC can be used to deal with some basic AI problems, such as search and pattern matching, as soon as quantum computers become widely available.

## 1. Introduction

*Quantum Computation* (QC) is the scientific field that studies how the quantum behavior of certain subatomic particles (i.e. photons, electrons, etc.) can be used to perform computation and eventually large scale information processing. *Superposition* and *entanglement* are two key-phenomena in the quantum domain that provide a much more efficient way to perform certain kinds of computations than classical algorithmic methods. In QC information is stored in *quantum registers* composed of series of *quantum bits* (or *qubits*). QC defines a set of operators called *quantum gates* that operate on quantum registers performing simple qubit-range computations. *Quantum algorithms* are successive applications of several quantum gates on a quantum register and perform more elaborate computations.

QC's ability to perform parallel information processing and rapid search over unordered sets of data promises significant advances to the whole scientific field of information processing. This article focuses on the benefits QC has to offer in the area of Artificial Intelligence (AI). In fact, several research papers have already reported how QC relates to specific aspects of AI (e.g. quantum game theory [Miakisz et al. (2006)], quantum evolutionary programming [Rylander et al. (2001)], etc). The present article attempts a more global view on quantum methods for AI applications addressing not only work already done but also some broad ideas for future work. But first it presents a very brief (due to space limitation) introduction to QC basics and algorithms, just the essentials to understand the subject. For a full introduction and more details the reader is advised to read [Karafyllidis (2005a)], [Gruska (1999)] or [Nielsen & Chuang (2000)].

## 2. Quantum Computation Basics

The quantum analog of a bit is called a *quantum bit* or *qubit*. Its physical implementation can be the energy state of an electron in an atom, the polarization of a photon, or any other bi-state quantum system. When a qubit is *measured* (or *observed*), its state is always found in one of two clearly distinct states, usually transcribed as |0> and |1>. These are direct analogs of the 0 and 1 states of a classical bit but they are also orthogonal states of a 2-dimensional Hilbert space and they are called *basis states* for the qubit. Before the qubit is measured, its state can be in a composition of its basis states denoted as:

$$|q\rangle = a|0\rangle + b|1\rangle = a\begin{bmatrix} 1 \\ 0 \end{bmatrix} + b\begin{bmatrix} 0 \\ 1 \end{bmatrix} = \begin{bmatrix} a \\ b \end{bmatrix} \qquad (1)$$

In Eq.1 a and b are complex numbers called *probability amplitudes*; $|a|^2$ is the probability of the qubit to appear in state |0> when observed, and $|b|^2$ is the probability to appear in state |1>. Equation 1 also presents the *matrix notation* of the qubit states. A series of qubits is called a *quantum register*.

An n-qubit quantum register is denoted as:

$$|Q_n\rangle = c_0|0\cdots000\rangle + c_1|0\cdots001\rangle + \cdots + c_{2^n-1}|1\cdots111\rangle = \sum_{i=0}^{2^n-1} c_i|i\rangle \qquad (2)$$

It has $2^n$ observable states, corresponding to the basis states of Eq.2, each one having a probability of $|c_i|^2$ when measured. Again, this can be considered as a vector of an n-dimensional Hilbert space with $\sum_{i=0}^{2^n-1}|c_i|^2 = 1$.

A single qubit can be considered as a trivial quantum register with n=1. When n>1 the quantum register can be considered as a series of qubits:

$$|Q_n\rangle = |q_{n-1}\rangle \otimes |q_{n-2}\rangle \cdots |q_i\rangle \cdots |q_1\rangle \otimes |q_0\rangle = |q_{n-1}q_{n-2}\cdots q_i \cdots q_1 q_0\rangle \qquad (3)$$

where $\otimes$ denotes the *tensor product*.

Quantum systems are able to simultaneously occupy different quantum states. This is known as a *superposition* of states. In fact, the state of Eq.1 for the qubit and the state of Eq.2 for the quantum register represent superpositions of the basis states over the same set of qubits. A quantum register can be in a superposition of two or more basis states (with a maximum of $2^n$, where n is the number of its qubits). The qubits of the quantum register remain in superposition until they are measured (intentionally or not). At the time of measurement the state of the register *collapses* (or *is resolved*) to one of its basis states randomly, according to the probability assigned to that state.

It is not necessary to measure every single qubit of a quantum register in order to trigger its collapse to a basis state. For example, consider this case:

$$\left|Q_5\right\rangle = \frac{1}{\sqrt{3}}\left|00000\right\rangle + \frac{1}{\sqrt{3}}\left|10000\right\rangle + \frac{1}{\sqrt{3}}\left|11111\right\rangle \qquad \textbf{(4)}$$

Equation 4 specifies a 5-qubit register in superposition of three (of the 32 possible) basis states, |00000>, |10000> and |11111> with equal probability amplitudes; each of the three states has a 33% chance to be observed. Now, suppose we measure the qubits one by one starting from the leftmost. The leftmost qubit has a 67% chance to be |1> and 33% to be |0>. Let's say we measure it and find a |0>. We say that the leftmost qubit has collapsed to |0>. But it is not the only qubit that has collapsed; the rest four qubits must be all |0> too, since these are the only states consistent with the leftmost |0>. We say that the four rightmost qubits are *entangled* with the leftmost one. In other words, they are linked together in a way that each of the qubits loses its individuality. Measurement of one affects the others as long as they remain entangled together. Note that if instead of measuring the leftmost qubit we had decided to measure the rightmost one and found it |0>, three other qubits would collapse to |0> as well, but the leftmost qubit would still remain in superposition. But this does not mean that it was not affected by the measurement; it now has a 50%-50% chance of being observed in |0> or |1> instead of the initial 33%-67%.

Superposition does not always imply entanglement. For example, consider the state of Eq.2: we have to measure each and every one of the n qubits in order to determine the exact state of the register. In this case there is no entanglement.

Quantum systems in superposition or entangled states are said to be *coherent*. This is a very fragile condition and can be easily disturbed by interaction with the environment (which is considered an act of measurement). Such an accidental disturbance is called *decoherence* and results to losing information to the environment. Keeping a quantum register coherent is very difficult, especially if its size is large.

## *3. Quantum Computation Components and Algorithms*

Higher order quantum computation machines can be devised based on quantum registers: for instance quantum finite state automata can be produced by extending probabilistic finite-state automata in the quantum domain. Analogous extensions can be performed for other similar state machines (e.g. quantum cellular automata, quantum Turing machines, etc) [Gruska (1999)]. Regardless the machine, the computation is eventually reduced to a series of basic operations to some qubits of a quantum register; this is what *quantum gates* do.

*Quantum gates* are the basic computation components for QC. They are very different from gates in classical computation systems. Quantum gates are not circuits with input and output; they are operators over a quantum register. These operators are always reversible; most of them originate from reversible computation theory.

An infinite number of quantum gates can be defined (even for a single qubit) since it is possible to define an operator that rotates an arbitrary quantum register state anywhere in the Hilbert space. The most common quantum gates are:

- *The Identity Gate:* It is the quantum equivalent of a buffer.
- *The NOT Gate:* It is used to complement the input.
- *The Hadamard Gate:* It is used to set a qubit in a superposition of two states. Acts on a single qubit.
- *The Phase Shift Gate:* In fact it is a class of gates with varying phases. It changes the phase of a qubit in the Hilbert space.
- *The Controlled NOT Gate (CNOT or XOR):* Like the NOT gate, but acts on two qubits. The first one is called *control qubit*, the second one *target*. The gate performs a complement of the target qubit only if the control qubit is |1>. This effect is equivalent to a XOR operation between the two qubits, hence the alternative name.
- *The Controlled Phase Shift Gate:* Like the Phase Shift gate, but acts on two qubits: control and target. It performs a phase shift on the target qubit only if the control qubit is |1>.
- *The Exchange Gate:* Acts on two qubits and exchanges their values.
- *The Controlled-Controlled NOT Gate (CCNOT or Toffoli):* Like the CNOT, but with two control qubits. Both of them should be |1> in order to complement the target qubit.
- *The Fredkin Gate:* Like the Exchange gate, with an additional control qubit. The two target qubits exchange their values only if the control qubit is |1>.

Each gate is expressed as a matrix, so that the application of a quantum gate on the contents of a quantum register is expressed as a matrix multiplication.

*Quantum Algorithms* are series of applications of quantum gates over the contents of a quantum register. The most popular quantum algorithms are:

- *Parallel Computation:* Thought not exactly an algorithm, the intrinsic property of quantum registers to support massively parallel computation is mentioned due to its use in almost every quantum algorithm. When a transformation is performed to the contents of a quantum register this affects the whole set of its superimposed values. Reading the outcome is a non-deterministic process, but it is possible to maximize the probability to occur the intended result. This is called *probability amplitude amplification* [Gruska (1999)].

- *Grover's Algorithm:* It searches $N=2^n$ items superimposed on a quantum register of n qubits using a certain state as a search key. It is able to search an unordered set of items in $O\left(\sqrt{N}\right)$ time [Grover (1997)].

- *Quantum Fourier Transform (QFT):* A basic subroutine in many specialized algorithms concerning factoring prime numbers and simulating actual quantum systems. QFT is a unitary operation acting on vectors in the Hilbert space. By altering their phases and probability amplitudes it can reveal periodicity in functions just like its classical analog [Coppersmith (1994)].

- *Shor's Algorithm:* It finds the period of a periodic function in polynomial time, a problem directly related to factorization of large integers [Shor (2004)]. This algorithm is famous for making obsolete the current public-key encryption systems.

## *4. On Quantum Artificial Intelligence*

One of the first contributions that QC offers to AI is the production of truly random numbers. True randomness has been reported to cause measurable performance improvement to genetic programming and other automatic program induction methods [Rylander et al. (2001)]. Monte-Carlo, simulated annealing, random walks and other analogous search methods are expected to benefit from that as well. A truly random number of N bits can be produced by applying the Hadamard transformation to a N-qubit quantum register thus producing the superposition of all basis states

$$\frac{1}{\sqrt{2^N}} \sum_{k=0}^{2^N-1} |k\rangle$$ . Then just by measuring this state we get a truly random number in the

range of $[0, 2^N-1]$. Since the process can be repeated n times to produce a nN-bit random number, it is generally possible to produce N-bit random numbers using a M-qubit quantum register where M<<N. Thus, in principle even just one qubit (M=1) is adequate.

However, random search methods in QC indicate a completely different approach than in classical computation. The quantum analog of a classical random walk on a graph, i.e. the quantum random walk, even in one dimension is a much more powerful computational model [Ben-Avraham et al. (2004)]. While the classical random walk is essentially a Markov process, in a quantum random walk propagation between node pairs is exponentially faster, thus enabling the solution of NP-complete problems as well [Childs et al. (2002)]. Moreover, as mentioned by [Shor (2004)], combinations of quantum random walks with Grover's algorithm have managed to confront efficiently some real-world problems like database element comparison and dense graph search [Childs et al. (2003)].

Grover's algorithm [Grover (1997)] and its variations are ideal for efficient content-addressable search and information retrieval from large collections of raw data. The principle of *probability amplitude amplification* that guides these processes can be relaxed for approximate pattern matching as well, thus facilitating applications like face, fingerprint, and voice recognition, corpus search, and data-mining. A quantum register containing a set of data in superposition can be seen as the quantum analog of a Hopfield neural network used as an associative memory [Trugenberger (2002)] only with much greater capacity to store patterns: while the capacity of a n-neuron Hopfield network approximates to 0.14n patterns, a quantum register of n-qubits can store $2^n$ binary patterns.

In principle, a great deal of the problems that AI attempts to confront is too heavy for classical algorithmic approaches, i.e. NP-hard problems such as scheduling, search, etc. Many AI techniques have been developed to cope with the NP-complete nature of these problems. Since QC can reduce time complexity to polynomial range, it eventually provides a more efficient way to address these problems. Using QC all the states of the search space can be first superimposed on a quantum register and then a search can be performed using a variance of Grover's algorithm. It is evident that many problems in search, planning, scheduling, game-playing, and other analogous fields can utilize the parallel processing of a quantum register's contents and reduce their processing times by several orders of magnitude. For more complex problems even quantum constraint satisfaction heuristics can be applied, as described in [Aoun & Tarifi (2004)]. But the main challenge in these cases is to find a way to encode the problem space within the quantum register boundaries. Fortunately, for problems where a previous approach based on genetic algorithms is available, there is a significant basis for QC as well: the representation of the gene-string can be transferred to the quantum implementation almost verbatim and the whole gene pool can be superimposed to a single quantum register.

Speech and language processing have also a great deal to gain from QC. Apart from the aforementioned approximate pattern matching to the input signal and the obvious rapid quantum search in huge lexical databases, the representation problem can be solved quite elegantly in a quantum register and more efficiently than ever. For instance, a common drawback of a typical syntactic parser is the fact that it produces too many parse trees, some slightly different and some quite different ones. Their representation as superimposed states in a quantum register solves not only the issue of their storage, but simplifies their further processing as well. An interesting model for the mapping of language expressions into microscopic physical states has been proposed by [Benioff (2002)].

Game theory and decision-making have also been addressed by QC. A new field of quantum game theory has emerged [Piotrowski & Sladkowski (2004a)] with promising applications at least to playing market games [Piotrowski & Sladkowski (2004b)]. The entanglement effect has been exploited to improve behavior in cooperation [Miakisz et al. (2006)] and coordination games [Huberman & Hogg (2003)], simulating economic systems [Chen et al. (2002)] and human behavior [Mendes (2005)]. It is interesting that in some cases the quantum solution can be derived as an extension to the classical one [Huberman & Hogg (2003)], thus enabling the optional use of quantum entanglement as an extra resource. Even more interesting is the fact that the quantum solution to these games models the human behavior much more accurately than the classical one. In fact, it seems that human player strategies in these types of games deviate significantly from the theoretical Nash equilibrium that classic game theory expects. This discrepancy (i.e. the seemingly irrational human behavior) is attributed to an emotional response of the human player. Quantum solutions such the one for the *Quantum Ultimatum Game* [Mendes (2005)] seem to model efficiently these discrepancies and propose a better model for the human behavior in such situations.

The last comment on modeling human behavior seems to lead to the philosophical question of whether the human brain performs some kind of quantum computation [Miakisz et al. (2006)] or not [Penrose (1997)], a question that has been used to argue against hard-AI in the past. Although there are not sufficient data to answer such a question, one could argue that with QC the idea of hard-AI seems a little closer to implementation, since there is some evidence that QC is stronger than classical Turing computation. Indeed, [Calude & Pavlov (2002)] have proved that QC is theoretically capable of computing incomputable functions. Despite Feynman's argument that QC is not able to exceed the so-called Turing's barrier and solve an undecidable problem, Calude & Pavlov have used QC to solve an equivalent to the famous Halting Problem, the most well-known undecidable problem in computer science. Although it still has to be seen whether this approach is practically feasible, this is a great theoretical breakthrough that promises to change the computational capabilities of our machines.

Eventually, QC's effectiveness may eventually make us reconsider what an AI problem is. For example chess playing is traditionally considered an AI problem. That is so due to the high computational cost of the brute-force algorithmic approach in the classical computational 'world'. But in the quantum domain the same problem is not so hard. Given the proper hardware, a quantum algorithmic process would be able to solve it in acceptable time. So does chess playing remain an AI problem? Only time will tell whether QC will force us to redefine the domain of AI or whether it will be eventually considered yet another weapon in the AI arsenal, like neural networks or genetic algorithms.

## 5. Conclusion

This paper attempted to present the basics of QC to readers already familiar with AI, explaining QC's potential application to traditional AI problems and methods through a very narrow choice of recent papers and research directions. For a lengthier overview on QC applications to Computational Intelligence see [Perkowski (2005)].

The ideas presented here were inevitably vague and outlined, since quantum computers are still not available for implementation purposes. The hardware for quantum registers is still in infancy due to the obstacle of decoherence which is very difficult to overcome. Thus most of the aforementioned methods have been tested either to trivial problems (requiring 3 to 5 qubits) or to QC simulators [Karafyllidis (2005b)]. But maybe this situation is going to change sooner than expected. Very recently the Canadian company D-Wave Systems announced a prototype 16-qubit adiavatic quantum computer that (among other things) solves Sudoku puzzles [Minkel (2007)]. The company has also promised to provide a commercial product very soon. Meanwhile, a programming language for quantum programming has already been proposed [Betteli et al. (2005)], so by the time quantum computers become available in the market, probably a great deal of software tools will be ready as well and the road to Quantum Artificial Intelligence will be open to explore.

## *References*

Aoun, B., Tarifi, M. (2004), *Quantum Artificial Intelligence*, Quantum Information Processing, ArXiv:quant-ph/0401124.

Ben-Avraham, D., Bollt, E.M., Tamon, C. (2004), *One-Dimensional Continuous-Time Random Walks*, Quantum Information Processing, vol.3, pp.295-308.

Benioff, P. (2002), *Language is Physical*, Quant.Infor.Processing, vol.1, pp.495-509.

Betteli, S., Calarco, T., Serafini, L. (2005), *Toward an Architecture for Quantum Programming*, ArXiv:cs.PL/0103009.

Calude, C.S., Pavlov, B. (2002), *Coins, Quantum Measurements, and Turing's Barrier*, Quantum Information Processing, vol.1, pp.107-127.

Chen, K.-Y., Hogg, T., Beausoleil, R. (2002), *A Quantum Treatment of Public Goods Economics*, Quantum Information Processing, vol.1, pp.449-469.

Childs., A.M., Cleve, R.E., Deotto, E. Farhi, E., Gutmann, S., Spielman, D.A.(2003), *Exponential Algorithmic Speedup by Quantum Walk*, in Proc. 35th ACM Symposium on Theory of Computing, pp.59-68.

Childs, A.M., Farhi, E., Gutmann, S. (2002), *An Example of the Difference Between Quantum and Classical Random Walks*, Quantum Inform. Proc., vol.1, pp.35-43.

Coppersmith, D. (1994), *An Approximate Fourier Transform Useful in Quantum Factoring*, IBM Research Report RC 19642.

Grover, L.K. (1997), *Quantum Mechanics Helps in Searching for a Needle in a Haysack*, Phys. Rev. Lett., vol.78, pp.325-378.

Gruska, J. (1999), *Quantum Computing*, McGraw-Hill, London.

Huberman, B.A., Hogg, T. (2003), *Quantum Solution of Coordination Problems*, Quantum Information Processing, vol.2, pp.421-432.

Karafyllidis, I.G. (2005a), *Quantum Computers – Basic Principles*, Klidarithmos, Athens (in Greek).

Karafyllidis, I.G. (2005b), *Quantum Computer Simulator Based on the Circuit Model of Quantum Computation*, IEEE Trans.Circ.& Syst.-I, vol.52, no.8, pp.1590-1596.

Mendes, R.V. (2005), *The Quantum Ultimatum Game*, Quantum Information Processing, vol.4, pp.1-12.

Miakisz, K., Piotrowski, E.W., Sladkowski, J. (2006), *Quantization of Games: Towards Quantum Artificial Intelligence*, Theor.Comp.Science, vol.358, pp.15-22.

Minkel, J. R. (2007), *First "Commercial" Quantum Computer Solves Sudoku Puzzles*, Scientific American News, February 13, 2007, http://www.sciam.com/article.cfm?articleID=BD4EFAA8-E7F2-99DF-372B272D3E271363

Nielsen, M.A., Chuang I.L. (2000), *Quantum Computation and Quantum Information*, Cambridge University Press, Cambridge.

Penrose, R. (ed.) (1997), *The Large, the Small and the Human Mind*, Cambridge University Press, Cambridge.

Perkowski, M.A. (2005), *Multiple-Valued Quantum Circuits and Research Challenges for Logic Design and Computational Intelligence Communities*, IEEE Connections, vol.3, no.4, pp.6-12.

Piotrowski, E.W., Sladkowski, J. (2004a), *The Next Stage: Quantum Game Theory*, in Mathematical Physics Frontiers, Nova Science Publishers Inc.

Piotrowski, E.W., Sladkowski, J. (2004b), *Quantum Computer: An Appliance for Playing Market Games*, Int. J. Quant. Information, vol.2, pp.495.

Rylander, B., Soule, T., Foster, J., Alves-Foss, J. (2001), *Quantum Evolutionary Programming*, in Spector, L. et al. (eds.), Proc. of the Genetic and Evolutionary Computation Conference (GECCO-2001), San Francisco, USA, pp.1005-1011.

Shor, P.W. (2004), *Progress in Quantum Algorithms*, Quantum Information Processing, vol.3, pp.5-13.

Trugenberger, C.A. (2002), *Quantum Pattern Recognition*, Quantum Information Processing, vol.1, pp.471-493.